\documentclass[11pt,letterpaper]{article}
\usepackage{emnlp2016}
\usepackage{times}
\usepackage{latexsym}
\usepackage{graphicx}
\usepackage{float}
\usepackage{dblfloatfix}
\usepackage[export]{adjustbox}
\usepackage{hyperref}

\emnlpfinalcopy

\def\emnlppaperid{***}

\title{Machine Comprehension Based on Learning to Rank}
\author{Tian Tian \and Yuezhang Li\\
Carnegie Mellon University\\
  {\tt ttian1, yuezhanl@andrew.cmu.edu}}

\date{}

\begin{document}

\maketitle

\begin{abstract}
Machine comprehension plays an essential role in NLP and has been widely explored with dataset like MCTest.~However, this dataset is too simple and too small for learning true reasoning abilities.~\cite{hermann2015teaching} therefore release a large scale news article dataset and propose a deep LSTM reader system for machine comprehension. However, the training process is expensive. We therefore try feature-engineered approach with semantics on the new dataset to see how traditional machine learning technique and semantics can help with machine comprehension. Meanwhile, our proposed L2R reader system achieves good performance with efficiency and less training data.
\end{abstract}


\begin{table*}[t]
\centering
\begin{tabular}{lllllll}
\hline
                & \multicolumn{3}{c}{\textbf{CNN}} & \multicolumn{3}{c}{\textbf{Daily Mail}} \\ \cline{2-7}
                & train      & dev      & test     & train        & valid       & test       \\ \hline
\# documents    & 90,266     & 1,220    & 1,093    & 196,961      & 12,148      & 10,397     \\
\# queries      & 380,298    & 3,924    & 3,198    & 879,450      & 64,835      & 53,182     \\
\# max entities & 527        & 187      & 396      & 371          & 232         & 245        \\
\# avg entities & 26.4       & 26.5     & 24.5     & 26.5         & 25.5        & 26.0       \\
\# avg tokens   & 762        & 763      & 716      & 813          & 774         & 780        \\
word count      & \multicolumn{3}{c}{118,497}      & \multicolumn{3}{c}{208,045}             \\ \hline
\end{tabular}
\caption{Dataset Statistics}
\label{tb:corpus}
\end{table*}

\section{Introduction}
Machine comprehension, as the central goal in NLP, has been explored with a variety of methods.~Based on the released dataset MCTest \cite{richardson2013mctest}, a lexical matching based method is proposed \cite{smithstrong}. It applies linguistic features to tackle this problem.~However, this method dives deep into the characteristics of the dataset, and may not generalize well to other datasets.~This problem also goes with the discourse relation model introduced in \cite{narasimhan2015machine} that explore causal, temporal and explanation relationships between two sentences, which does not scale well.~Hence, \cite{wang2015machine} introduced a max-margin learning frame that incorporates a feature set of frame semantics, syntax, coreference, and word embeddings.~The accuracy of $69.94\%$ was achieved in MC500 of MCTest dataset.~Meanwhile, \cite{sachan2015learning} applied similar loss function, modeling machine comprehension as textual entailment and solved the problem by constructing latent answer-entailing structure with an accuracy of $67.83\%$.~Although such good accuracy were achieved, the dataset they worked on has some limitations in terms of data size and content.~Therefore, our work is based on a much larger news article dataset created by~\cite{hermann2015teaching}.

Different from the MCTest dataset, this news article dataset consists of cloze style questions, i.e.~questions generally generated by removing a phrase from a sentence \cite{taylor1953cloze}. Since the questions can be formed from a short summary of the document with condensed form of paraphrase, the dataset is suitable for testing machine comprehension \cite{hermann2015teaching}.

To realize machine comprehension, we develop a learning to rank reader (L2R Reader) system by first exploring features on frequency, word distance, syntax and semantics. Then, through learning to rank, we construct a ranking model that can directly pick the answer from the candidate list of answers. Opposed to the deep LSTM \cite{hermann2015teaching} that computes the answer based on context information of documents, it is more efficient and does not require much data to reach good performance. Moreover, we incorporate the semantics into the system to improve comprehension ability.

This article is organized as follows.~Section~\ref{sec:td} introduces the task and essential parts of relevant datasets. Section~\ref{sec:rw} presents related work and distinguishes our work from them. Section~\ref{sec:model} describes our model from aspects of learning to rank algorithms, features and usage of semantic information. Section~\ref{sec:evaluation} then evaluates our model from performance, semantics analysis and error analysis. Finally, section~\ref{sec:conclusion} summarizes our work and points out contributions.

\begin{figure*}[h]
    \centering
    \includegraphics[width=\textwidth]{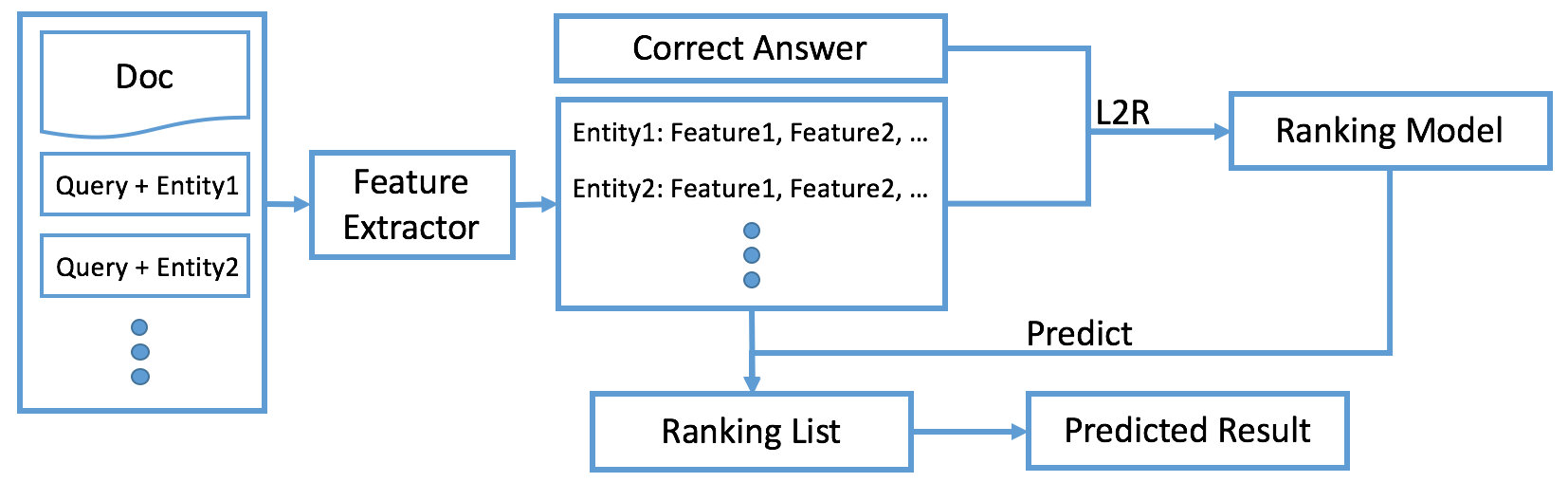}
    \caption{A high-level architecture of the L2R reader system}
    \label{fig:architecture}
\end{figure*}
\section{Task and Datasets}
\label{sec:td}
This section gives a brief introduction of the task and the dataset recently released for this task.
\subsection{Formal Task Description}
This task requires answering a cloze style question based on the understanding of a context document provided with the question. Along with each question and document, it also provides the correct answer to the question and a list of candidate answers. Thus, this can be formalized as follows:

The training data consists of tuples $(d, q, a, A)$, where $d$ is a context document for answering the question $q$, $a$ is the correct answer to question $q$, $A$ denotes a set of candidate answers to the question and $a \in A$ as defined.
 
\subsection{Datasets}
The dataset \cite{hermann2015teaching} we used in this task were constructed from news article from CNN and Daily Mail websites. The context document of the dataset is from the main body of the news article while the question is formed from one top sentence summarized the news article. Specifically, the question is constructed by replacing the named entity with a placeholder, e.g.~``@placeholder and @entity2 welcome son @entity6" is a question defined in the dataset. 

\begin{table}[h]
	\centering
	\includegraphics[width=0.4\textwidth]{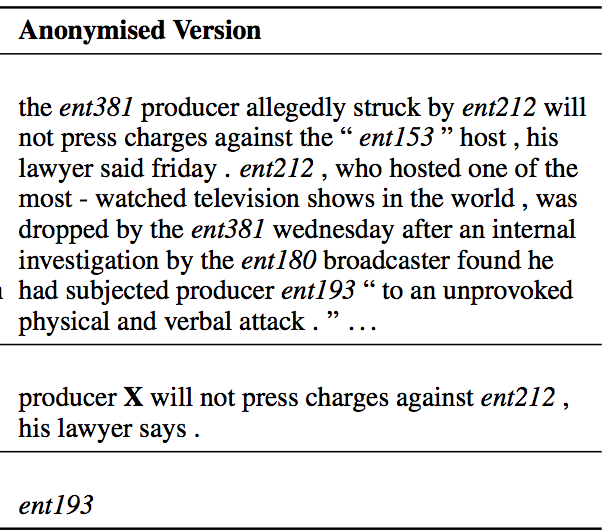}
	\caption{Example of Anonymised version of a data point}
	\label{fig:text}
\end{table}

Furthermore, to make context information required for answering question, we use the anonymised version as shown in Figure~\ref{fig:text} to eliminate influence of background knowledge. Thus, we must exploit the context to answer the question and these two corpora truly measures the capacity of reading comprehension. 

The basic statistics of CNN and Daily Mail of dataset are summarized in Table~\ref{tb:corpus}.

\section{Related Work}
\label{sec:rw}
Machine comprehension generally concentrates on MCTest \cite{richardson2013mctest} and due to the limitation of data size, the state of the arts are mainly based on traditional machine learning techniques. For example, \cite{wang2015machine} proposed a max-margin learning framework that combines features on syntax, coreference, frame semantics and word embeddings, which achieves significant improvement on the problem of MCTest question answering. Although recently \cite{trischler2016parallel} proposed a parallel-hierarchical model based on neural network and this method outperforms the previous feature-engineered approaches, it has reasoning limitation that the reasoning can only be achieved by stringing important sentences together. Their experiment proves that MCTest is too simple to learn true reasoning and it is also too small for that goal.

Considering the limitations of MCTest dataset, \cite{hermann2015teaching} provides a large scale supervised reading comprehension dataset collected from the CNN and Daily Mail websites. This helps with the bottleneck that large dataset is missing on machine comprehension evaluation. With this dataset,  \cite{hermann2015teaching} propose a deep LSTM reader that achieves an accuracy of $63.8\%$ on CNN and $69.0\%$ on Daily Mail. However, this deep LSTM reader is time consuming for training and no explanation can be found on why it works opposed to traditional approach. Therefore, we propose a traditional machine learning method on this new dataset to investigate what features can help with this task.

\section{Model -- Learning to Rank (L2R) Reader}
\label{sec:model}
Our model consists of learning to rank algorithms, features and semantics. As shown in Figure~\ref{fig:architecture}, given a document along with a list of queries that have the placeholder filled with different entities from the list of candidate answers. Then, through a feature extractor, we get several features for each entity (i.e.~candidate answer). Combining with correct answer, we employ a learning to rank algorithm (i.e.~L2R in Figure~\ref{fig:architecture}) to train a ranking model. Based on the ranking model, we generate ranking lists on new unseen dataset that has features extracted using feature extractor; from the rank list, we select the entity with highest ranking score as the predicted answer to the question.

\subsection{Learning to Rank}
Learning to rank is employed in Information Retrieval (IR) and Natural Language Processing (NLP). Generally, it is defined as follows: given a query, the ranking algorithm will generate a list of candidate documents with scores \cite{hang2011short}. In this task, we select the entity with the highest score as the answer to the question. Here, we introduce three different types of learning to rank algorithms that can help with the task. 

	\textbf{Pointwise}: In the pointwise approach, the ranking algorithm is transformed into problems including classification and regression to derive a score for every pair of document and query \cite{hang2011short}. This approach ignores the group structure of ranking and we do not apply it in this task.

	\textbf{Pairwise}: In the pairewise approach, the ranking algorithm is transformed into problems of pairewise classification or pairwise regression \cite{hang2011short}. It also ignores the group structure of ranking. In this project, we mainly focus on the pairewise classification that employs a binary classifier in document pair ranking. We try approaches including RankNet \cite{burges2005learning}, RankBoost \cite{freund2003efficient}, RankSVM \cite{herbrich1999large}, MART and LambdaMART \cite{burges2010ranknet}
	 
	\textbf{Listwise}: In the listwise approach, the ranking problem is addressed by taking the ranking list as instances in both learning and prediction process \cite{hang2011short}. It maintains the group structure and we employ the following listwise approaches: ListNet \cite{cao2007learning}, AdaRank \cite{xu2007adarank}, Coordinate Ascent.

\subsection{Features}
\label{sub:features}
We explore four types of features in this task. We start with the frequency of entity in both document and cloze-style question. Then we try features on word distance with different settings of window size. We further investigate features on syntactics and semantics to see how these affect the model performance.

\subsubsection{Frequency}
Frequency is explored based on one baseline of \cite{hermann2015teaching}. We simply count the number of entity that is in the candidate answer list appearing in the document and the question, then the count number works as the frequency feature of the entity. If this entity does not show in question or document, we assign a value of $0$.~The idea behind is that news article usually mentions important entities multiple times and cloze-style question is concerned with such entities.

\subsubsection{Word Distance}
\label{subsubsec:WD}
We investigate word distance from three aspects -- word alignment, nBOW, and word mover's distance (WMD) \cite{kusner2015word}. 

\begin{figure}[h]
		\includegraphics[width=0.48\textwidth, right]{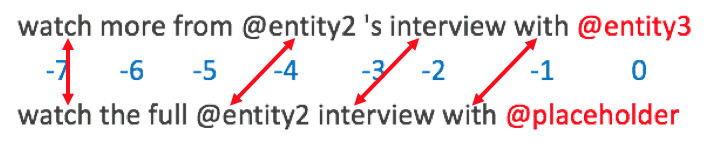}
		\caption{Example of Word Alignment}
		\label{fig:wa}
	\end{figure}

	\textbf{Word Alignment (WA)}: For the word alignment, we first consider the situation shown in Figure~\ref{fig:wa}.
	According to Figure~\ref{fig:wa}, we first replace the ``@placeholder" with one entity in the candidate answer list and search the document to find one matched sentence that containing this entity.~Then, we align these two entities and set location index as $0$. Starting with this index, words located left will have negative index while words located right will have positive index. With index defined, we align same words of these two sentences and compute the difference of word indexes. As for words without alignment, a penalty is given. Finally, the score capture some word information of question and document sentences.
	
	\textbf{Normalized Bag-Of-Words (nBOW)}: Furthermore, we consider using normalized bag-of-words (nBOW) vectors, $d \in R^n$, where $d$ denotes the document vector and $n$ is the vocabulary size. To be precise, if word $i$ appears $c_i$ times in the document, we let $d_i = \frac{c_i}{\sum_{j=1}^n c_j}$. Therefore, a sentence is transformed into a vector and we can do similarity measure accordingly.
	
\begin{figure}[h]
		\includegraphics[width=0.45\textwidth, right]{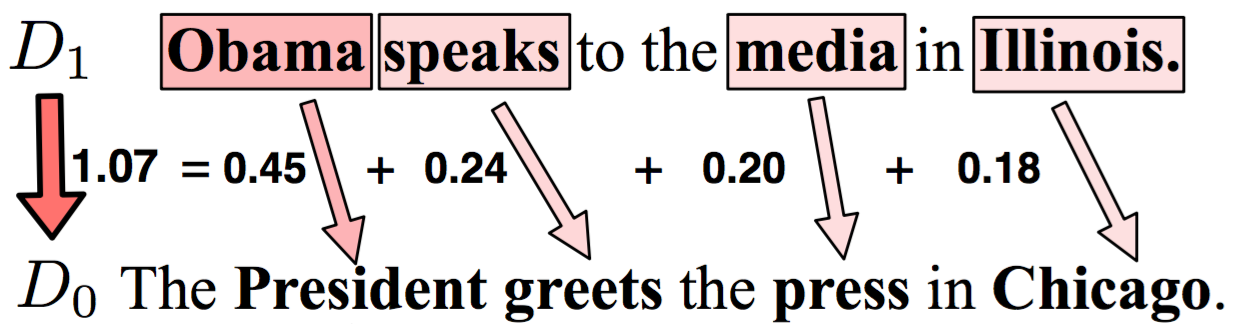}
		\caption{Example of WMD}
		\label{fig:wmd}
	\end{figure}

	\textbf{Word Mover's distance (WMD)}: The WMD \cite{kusner2015word} measures the dissimilarity between two text documents using the minimal distance that the embedded words of one document need to move to the embedded words of another document.~Here, we apply WMD into two sentences after using Mikolov's word2vec \cite{mikolov2013distributed} to convert words of sentences to embeddings. Different from the two methods above (WD and nBOW), the WMD can move words to semantically similar words, thereby capturing semantic information (see Figure~\ref{fig:wmd}). It can capture semantic similarity of two sentences with different unique words. For instance, ``The President greets the press in Chicago" and ``Obama speaks to the medis in Illinoirs".
	
	Moreover, we try different window size to extract sentences in the document to optimize our model.

\subsubsection{Syntactic Features}
In this task, we consider syntactic features including part-of-speech (POS) tags and dependency parsing.

\textbf{POS tags}: Similar to the word alignment defined in Section~\ref{subsubsec:WD}, we consider POS tag alignment as one syntactic feature. Specifically, we transform words into POS tags using NLTK\footnote{\url{http://www.nltk.org/}} and employ same technology as word alignment to measure the dissimilarity of two sentences. 

\textbf{Dependency Parsing}: If two sentences describe same event, it is likely that they have dependency overlapping \cite{wang2015machine}. We thus incorporate dependency parsing to capture such information. To be precise, we use Stanford Parsing\footnote{\url{http://nlp.stanford.edu/software/stanford-dependencies.shtml}} to get dependencies of a sentence that are shown as several triples like ($s$, $t$, $arc$), e.g.~(entity, has, nsubj), where $s$ denotes source word and $t$ is the target word. Then, this dependency-based similarity is evaluated from these three categories: (1) $s_d$ = $s_q$ and $arc_d = arc_q$ or $t_d$ = $t_q$ and $arc_d = arc_q$; (2) $s_d$ = $s_q$ and$t_d$ = $t_q$; (3) $s_d$ = $s_q$, $t_d$ = $t_q$ and $arc_d = arc_q$, where $d$ denotes document, $q$ denotes question and $s_d$ refers to source word in document sentence while $s_q$ represents the target word in question.

\subsubsection{Semantic Features}
\begin{figure}[h]
    \includegraphics[width=0.5\textwidth]{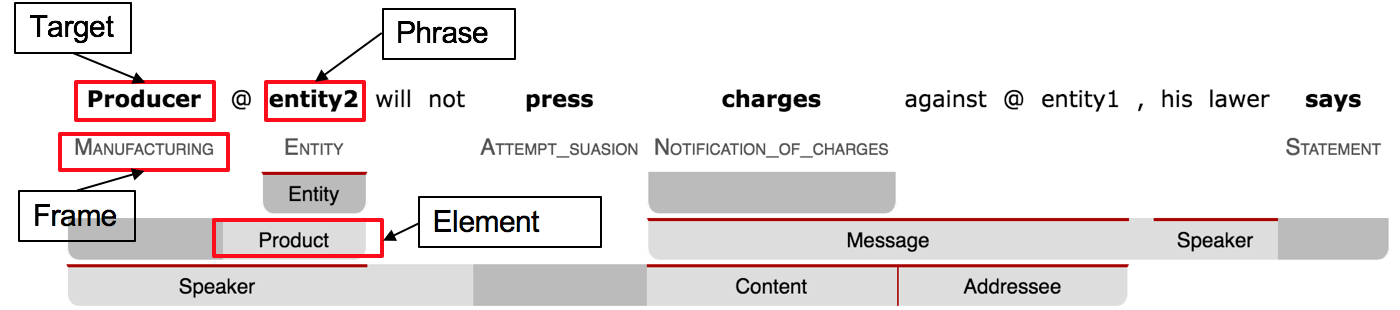}
    \caption{Example output from SEMAFOR}
    \label{SEMAFOR}
\end{figure}

In addition to word embeddings applied in WMD in Section~\ref{subsubsec:WD}. We adopt the SEMAFOR Frame Semantic (FS) parser \cite{das2014frame} to extract some semantic features. Figure~\ref{SEMAFOR} gives an example output of the SEMAFOR semantic parser. 
In this example, five frames are identified. For example, The word ``says" is a \textbf{target}, which evokes a semantic \textbf{frame} labeled \textsc{STATEMENT}. Each frame has its own frame \textbf{elements}; e.g., the \textsc{STATEMENT} frame has frame elements of \textit{Message} and \textit{Speaker}. Features from these parsers have been shown to be useful for machine comprehension task~\cite{wang2015machine}. We expect that the document sentence containing the answer will overlap with the question and correct answer in terms of targets, frames evoked and frame elements evoked. Therefore, we design the following features to capture this intuition. To be precise, after parsing a sentence, we get several triples composed of $(t, f, e)$, where $t$ denotes the target, $f$ denotes the frame and $e$ denote a set of elements. Then, the frame semantic based features are derived from the next seven categories: (1) $t_q = t_d$ ; (2) $f_q = f_d$; (3) $e_q = e_d$; (4) $t_q = t_d$ and $f_q = f_d$ ; (5)$t_q = t_d$ and $e_q = e_d$; (6) $f_q = f_d$ and $e_q = e_d$; (7) $t_q = t_d$ and $f_q = f_d$ and $e_q = e_d$. We count the number of triples satisfying the above requirements from the document sentence and the query sentence to generate seven features.

\subsection{Semantics}
Semantics play a significant role in our model.~This section summarizes how our model uses semantics to achieve reading comprehension. In this project, we use semantics in aspects of WMD, Frame Semantic (FS) and coreference.~In WMD, we use word embeddings to capture word level semantics and FS helps us to capture sentence level semantics. The coreference is employed to identity chains of mentions within and across sentences for data pre-processing. 

Specifically, word embeddings \cite{mikolov2013distributed} project words into a low-dimensional space and similarity of vectors can capture some word similarity on semantics.~For example, the word ``Paris" is close to ``Berlin" rather than ``France" and vec(``Paris") is closest to vec(``Berlin") - vec(``Germany") + vec(``France"), where vec($x$) denotes the word embedding of word $x$. Based on the word embedding given by \cite{mikolov2013distributed}, the WMD can align semantically similar words together using distance measure, for example, it is much cheaper to transform ``Illinois" into ``Chicago" than ``Japan" into ``Chicago". Therefore, we incorporate semantics into the model and its performance is improved.

As for frame semantics, the semantic similarity that two sentences describe same events but use different words or two sentence have different structures can be captured by frame semantics parsing. For example, two sentence ``the speaker states that he is innocent." and ```I'm innocent', he says." would be parsed to have the same semantic frame of STATEMENT and frame elements of (Message, Speaker), even they use different words and have inverse sentence order.

Coreference resolution is achieved using Stanford CoreNLP\footnote{\url{http://stanfordnlp.github.io/CoreNLP/}}. We try to resolve the pronoun with the specific description and run the coreference resolution system on each document. As the coreference system will provide a chain of mentions, we take the representative mention to resolve pronoun in the document only, i.e.,~replacing the pronoun like ``it" with the representative mention of its coreference chain.

\section{Evaluation}
We evaluate the L2R reader system by comparing with the baselines of \cite{hermann2015teaching}.~Moreover, we present the system performance with different learning to rank algorithms, based on which, we select RankSVM and LamdaMART as the ranking algorithm for model training. We also evaluate the contribution of single feature to the system performance for final feature decision. Based on the final results, we analyze the effect of incorporating semantics into the system from coreference, word embeddings, and frame semantics. Following by semantic analysis, we conduct error analysis to see how to improve the system in the future work.

\label{sec:evaluation}
\subsection{Experimental Results}
\textbf{Final Results}: Table~\ref{result} shows the best performance of our model and some baselines. Our L2R model finally combines all the three kinds of word distance features and frame semantic features described in section~{\ref{sub:features}}. Our \textbf{L2R Reader} outperforms LSTM-based models proposed in \cite{hermann2015teaching} on the CNN dataset and achieves competitive results on the Daily Mail dataset. 
\begin{table}[h]
\centering
\begin{tabular}{|c|c|c|c|c|}
\hline
                 & \multicolumn{2}{c|}{CNN}      & \multicolumn{2}{c|}{Daily Mail} \\ \hline
                 & Train         & Test          & Train           & Test          \\ \hline
Deep LSTM$^\dagger$        & 55            & 57            & 63.3            & 62.2          \\ \hline
Attentive Reader$^\dagger$ & 61.6          & 63            & \textbf{70.5}   & \textbf{69}   \\ \hline
Impatient Reader$^\dagger$ & 61.8          & 63.8          & 69              & 68            \\ \hline
L2R Reader        & \textbf{64.3} & \textbf{65.8} & 69.1            & 67.3          \\ \hline
\end{tabular}
\label{result}
\caption{Results of our L2R Reader on the CNN and Daily Mail dataset. Results marked with $^\dagger$ are taken from previous paper.}
\end{table}

\textbf{Different L2R Algorithms}: 
Table~\ref{L2R} shows the performance of different learning to rank algorithms by using the same features.~Model parameters are tuned on the validation set. We found that \textbf{RankSVM} and \textbf{LambdaMART} are two best L2R algorithms on this task. RankSVM performs best on the CNN dataset while LambdaMART performs best on the Daily Mail dataset.

\begin{table}[h]
\centering
\begin{tabular}{|c|c|c|}
\hline
\begin{tabular}[c]{@{}c@{}}Ranking\\ Model\end{tabular}       & \begin{tabular}[c]{@{}c@{}}CNN\\ Test\end{tabular} & \begin{tabular}[c]{@{}c@{}}Daily Mail \\ Test\end{tabular} \\ \hline
RankSVM                                                       & \textbf{65.8}                                               & 66.7                                                       \\ \hline
MART                                                          & 60.4                                               & 65.3                                                       \\ \hline
RankNet                                                       & 40.9                                               & 32.8                                                       \\ \hline
RankBoost                                                     & 32.0                                               & 28.4                                                       \\ \hline
AdaRank                                                       & 18.0                                                & 12.7                                                        \\ \hline
\begin{tabular}[c]{@{}c@{}}Coordinate \\ Asecent\end{tabular} & 59.0                                               & 54.4                                                       \\ \hline
LambdaMART                                                    & 64.2                                               & \textbf{67.3}                                                       \\ \hline
ListNet                                                       & 32.7                                               & 32.3                                                       \\ \hline
\begin{tabular}[c]{@{}c@{}}Random \\ Forest\end{tabular}      & 63.4                                               & 65.6                                                       \\ \hline
\end{tabular}
\caption{Performance of different L2R algorithms on CNN and Daily Mail dataset.}
\label{L2R}
\end{table}

\textbf{Single Feature Performance}: We evaluate the performance of each single feature.~Table~\ref{single feature} shows the results. We can see that among all single features, word alignment features perform best and frame semantic features perform worst. This can be explained by the nature of the dataset.
\begin{table}[h]
\centering
\begin{tabular}{|l|l|l|l|l|}
\hline
                & \multicolumn{2}{l|}{CNN} & \multicolumn{2}{l|}{Daily Mail} \\ \hline
                & Train       & Test       & Train          & Test           \\ \hline
Frequency       & 33.2       & 35.0      & 33.4          & 32.4          \\ \hline
WA              & 47.3       & 50.6      & 54.3          & 53.4          \\ \hline
nBOW            & 40.3       & 43.5      & 48.7          & 47.8          \\ \hline
WMD             & 41.2       & 44.0      & 49.3          & 48.5          \\ \hline
FS & 18.9       & 22.0      & 25.2          & 24.3          \\ \hline
\end{tabular}
\caption{Single feature performance on CNN and Daily Mail dataset.}
\label{single feature}
\end{table}

\begin{figure}[h]
	\includegraphics[width=0.48\textwidth, right]{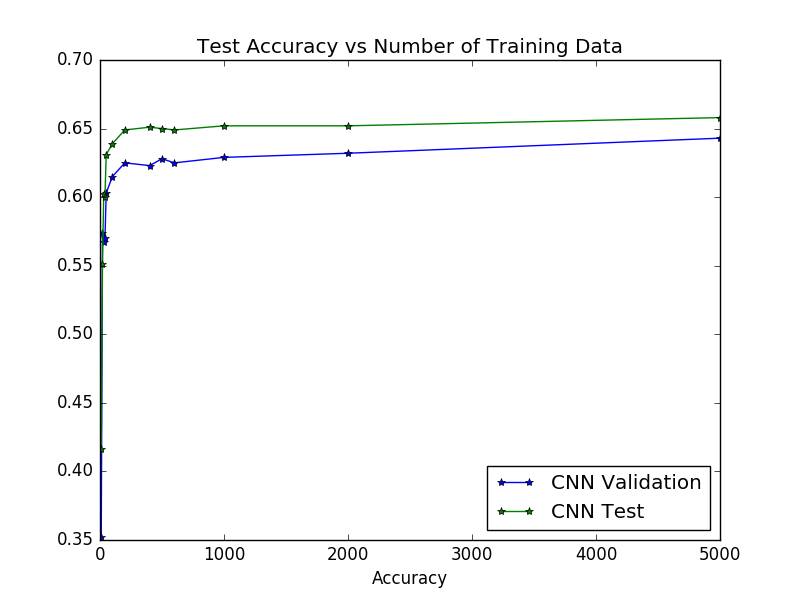}
	\caption{Model performance for L2R Reader against different number of training data on the CNN dataset.}
	\label{fig:performance vs data size}
\end{figure}

\textbf{Performance vs Size of Training Data:}
One great advantage of our model is that it only requires a small set of training data compared with neural network based models.~Table~\ref{performance vs data size table} shows the relationship between our model performance and the number of training data. We can see that our model have a high score given only about 100 training data. Our model performance improves when giving more data, but the speed of improvement gets dramatically slow. Figure~\ref{fig:performance vs data size} shows the trend of the convergence of our model with more data.

\begin{table}[h]
\centering
\begin{tabular}{|c|c|c|c|c|}
\hline
                & \multicolumn{2}{l|}{CNN} & \multicolumn{2}{l|}{Dailymail} \\ \hline
\#Training & Val     & Test    & Val        & Test       \\ \hline
10              & 35.2           & 41.6    & 45.7              & 43.6       \\ \hline
20              & 57.4           & 55.1    & 57.2              & 56.2       \\ \hline
30              & 56.7           & 60.2    & 61.4              & 60.3       \\ \hline
40              & 57             & 60      & 62.3              & 61.3       \\ \hline
50              & 60.3           & 63.1    & 63.5              & 62.5       \\ \hline
100             & 61.5           & 63.9    & 65.4              & 64.6       \\ \hline
200             & 62.5           & 64.9    & 67.3              & 65.2       \\ \hline
500             & 62.8           & 65      & 67.5              & 66.3       \\ \hline
1000            & 62.9           & 65.2    & 68.3              & 66.7       \\ \hline
2000            & 63.2           & 65.2    & 69.0              & 67.3       \\ \hline
5000            & 64.3           & 65.8    & 69.1                & 67.3       \\ \hline
\end{tabular}
\caption{Model performance for L2R Reader against different number of training data on the CNN and Daily Mail dataset. All the statistics are averaged over 10 times of random samples. Parameters are tuned on the validation set and the best validation model is applied to the test data.}
\label{performance vs data size table}
\end{table}

\subsection{Semantics Analysis}
We also test the performance of our semantic components, which are \textbf{co-reference}, \textbf{word embeddings} and \textbf{frame semantics}. 

\begin{table}[h]
\centering
\begin{tabular}{|c|c|c|c|c|}
\hline
           & \multicolumn{2}{c|}{CNN} & \multicolumn{2}{c|}{Daily Mail} \\ \hline
Models     & Val     & Test    & Val        & Test       \\ \hline
L2R Reader & 64.3           & 65.8    & 69.1              & 67.3       \\ \hline
L2R+Coref  & 63.8           & 64.8    & 68.3              & 66.5       \\ \hline
L2R-WMD    & 60.8           & 61.5    & 63.2              & 61.6       \\ \hline
L2R-FS     & 61.5           & 62.5    & 65.3              & 63.7       \\ \hline
\end{tabular}
\caption{Analysis of semantic components of our model. ``+"" and ``-" refer to added or ablated components. Coref, WMD and FS denotes coreference system, word mover's distance and frame semantics.}
\label{semantics}
\end{table}

Table~\ref{semantics} shows the results of adding co-reference, deleting word mover's distance and deleting frame semantics. From the experimental results, we can see that coreference system does not bring improvement to our system, and even harm the performance. This might because the coreference system (Stanford CoNLP) we use in our system cannot have good performance when the sentences are complicated. We can also find from the results that the \textbf{word embedding} and \textbf{frame semantics} components plays a vital role in our system, although the single feature performance of frame semantics is quite low.

\subsection{Error Analysis}
\begin{figure}[h]
	\includegraphics[width=0.48\textwidth, right]{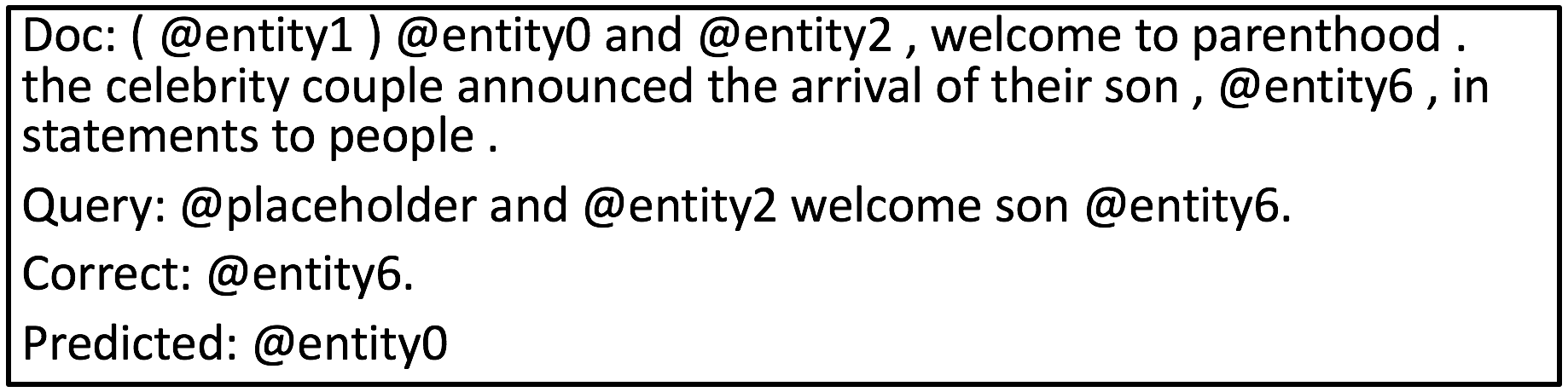}
	\caption{Wrong answer in the gold standard data.}
	\label{fig:Ex1}
\end{figure}

To give insight into our system's performance and reveal future research directions, we analyze the errors made by our system. We found that many queries require text summarization, event detection, background knowledge and inference. We also found an error on the CNN dataset (refer to Figure~\ref{fig:Ex1}). We make some detailed analysis as follows.

Figure~\ref{fig:Ex1} shows an example of wrong answer in the gold standard dataset. We can see from the query that the correct answer should be entity0 rather than entity6. Our model successfully get the right answer. The error in the dataset might due to the generation process of the dataset.

\begin{figure}[h]
	\includegraphics[width=0.48\textwidth, right]{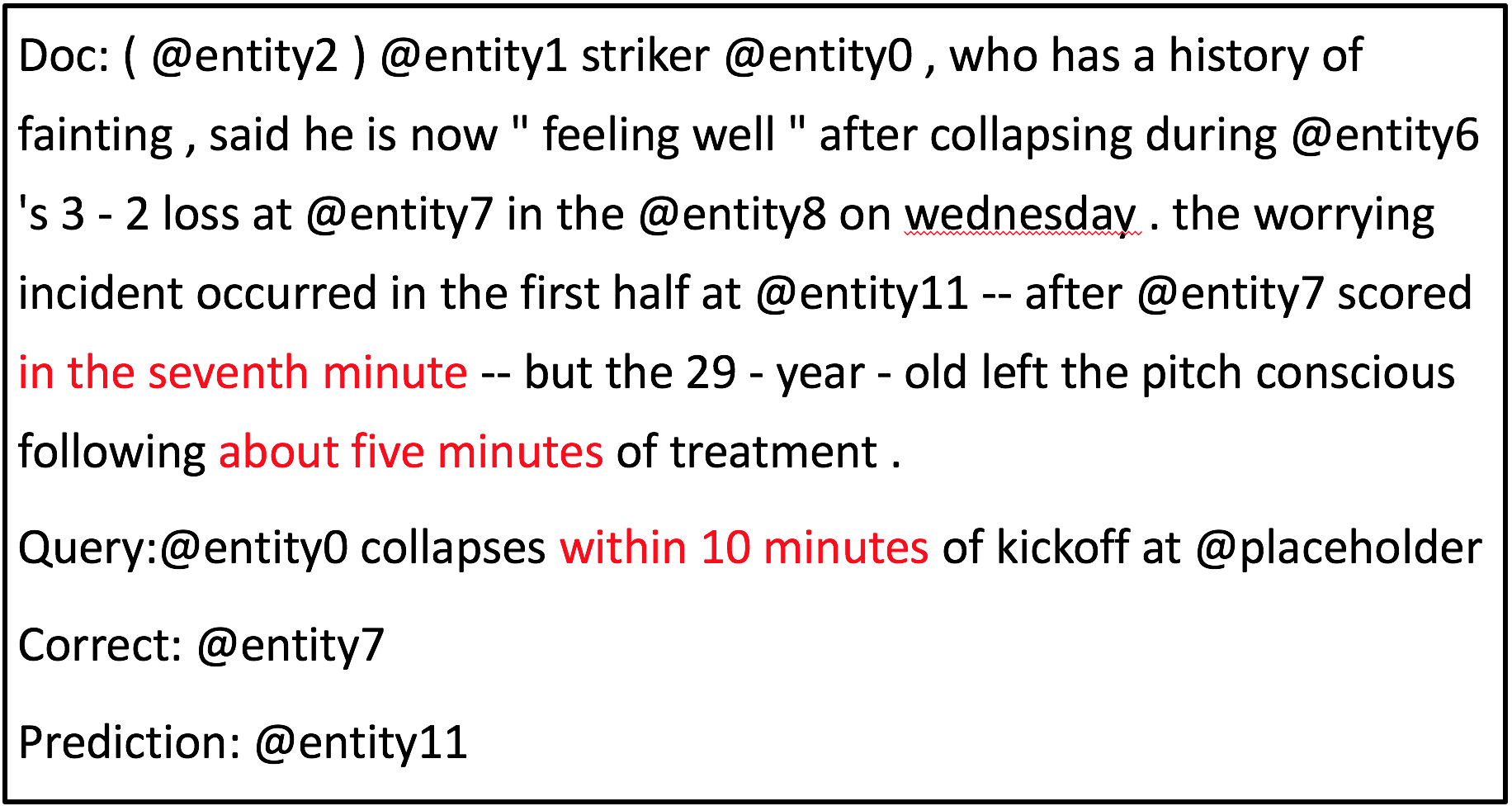}
	\caption{Require high level text summarization.}
	\label{fig:Ex2}
\end{figure}

Figure~\ref{fig:Ex2} shows an example of requiring high level text summarization. The phrases ``within 10 minutes" and ``kick off" do not appear in the document, but they are high level summarization of the document. Hence, our model lacks the ability of getting the correct answer in such situation. 

\begin{figure}[h]
	\includegraphics[width=0.48\textwidth, right]{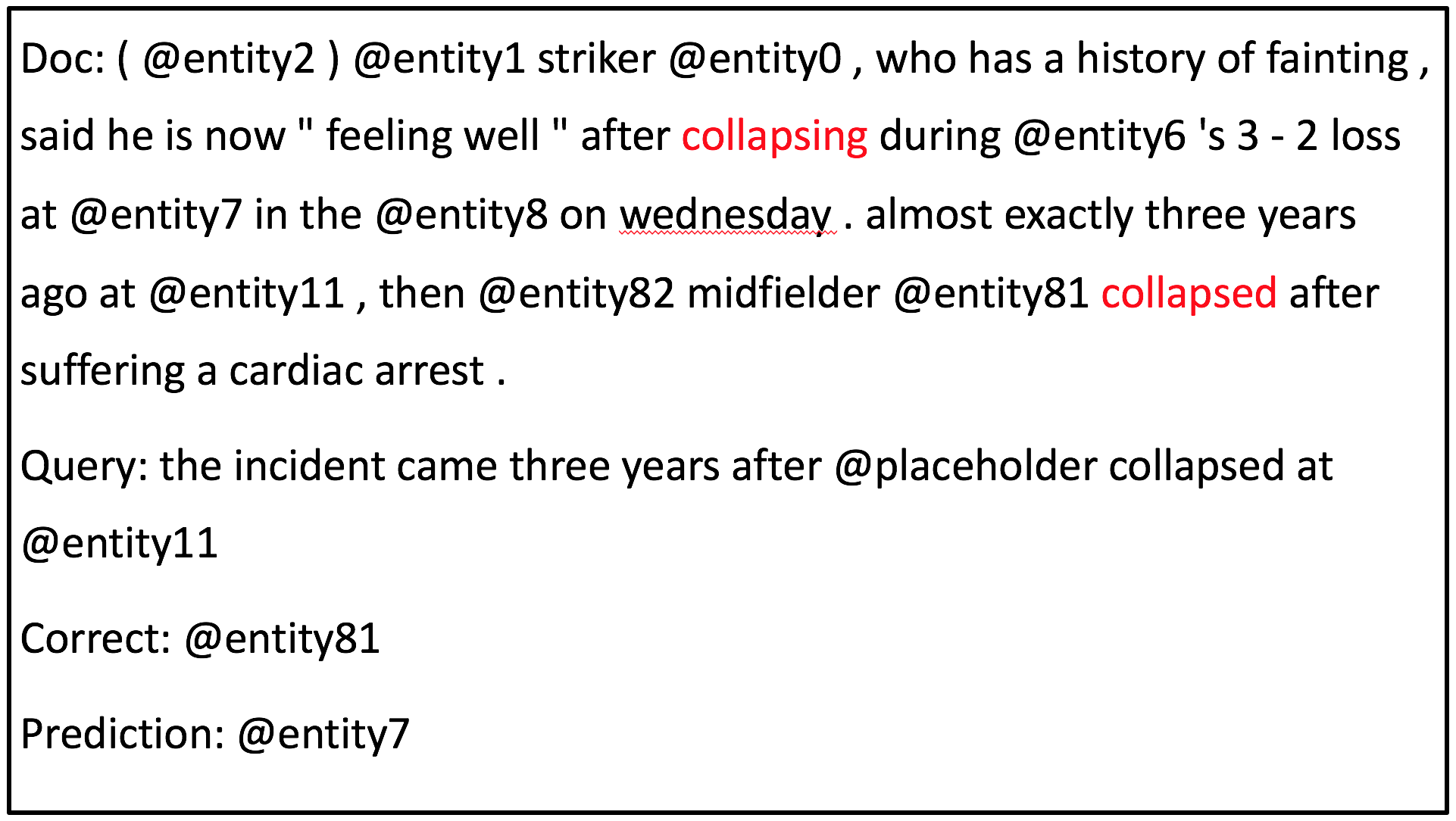}
	\caption{Require event detection.}
	\label{fig:Ex3}
\end{figure}

Figure~\ref{fig:Ex3} shows an example of requiring event detection in filling the cloze question. The word ``collapse" appears several times in the document, but describes different events. Our model fails to capture the difference. 

\begin{figure}[h]
	\includegraphics[width=0.48\textwidth, right]{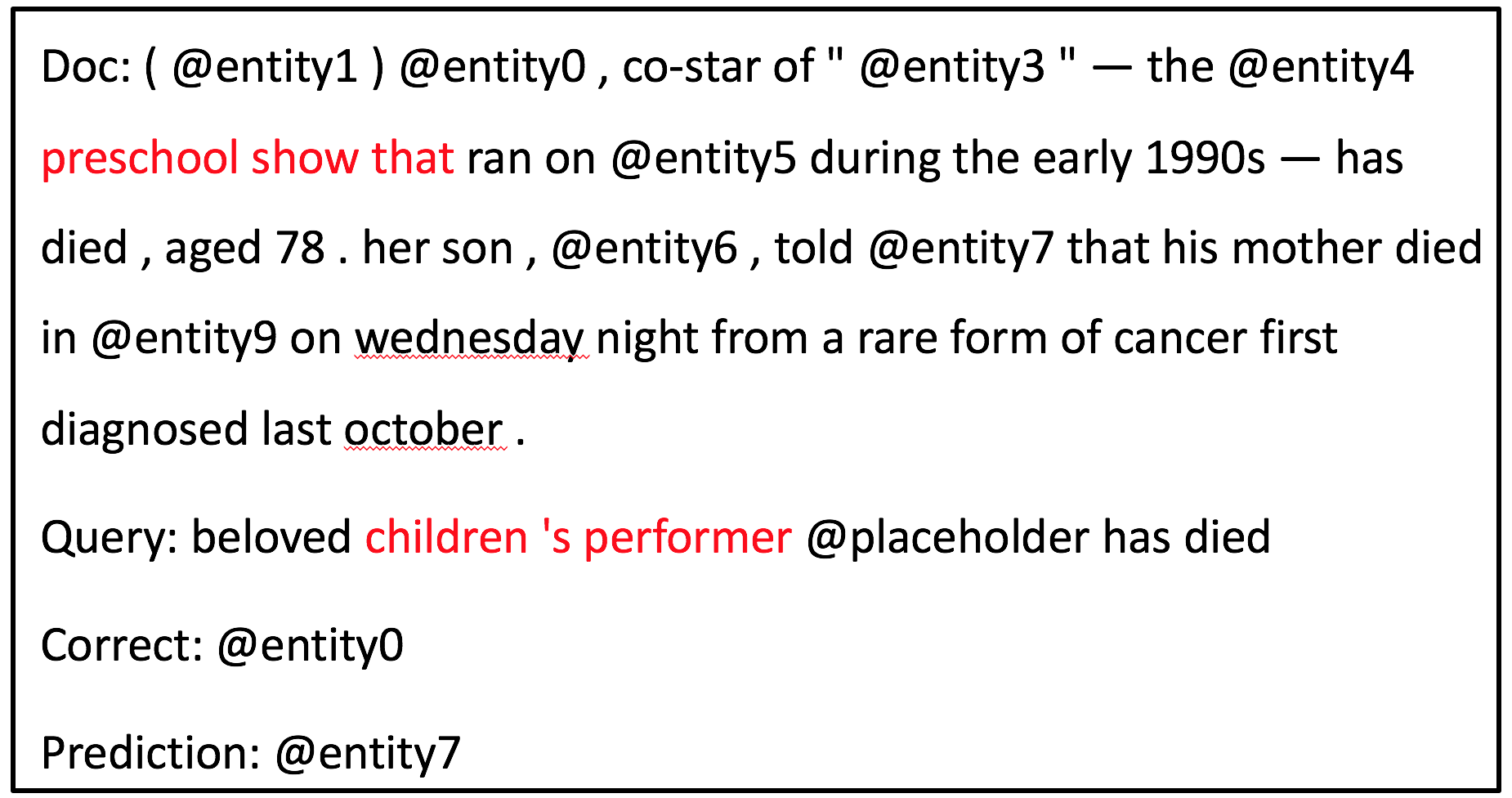}
	\caption{Require background knowledge and inference.}
	\label{fig:Ex4}
\end{figure}

Figure~\ref{fig:Ex4} shows an example of requiring background knowledge and inference in filling the cloze question. It can be inferred that ``preschool show" is performed by ``children's performer". However, this inference require some background knowledge. Our model cannot perform well in such situation.

\section{Conclusion}
\label{sec:conclusion}
We explored the new cloze style reading comprehension task and designed a learning to rank (L2R) reader system to provide a solution. We incorporate semantics such as word embeddings, frame semantics and coreference resolution into our system and show that they can greatly improve our model performance. We find that our model is poor at high level text summarization, event detection and inference through error analysis. We will investigate into how to solve these kinds of problems by using semantics in the future.

\bibliography{emnlp2016}

\begin{thebibliography}{}

\bibitem[\protect\citename{Burges \bgroup et al.\egroup
  }2005]{burges2005learning}
Chris Burges, Tal Shaked, Erin Renshaw, Ari Lazier, Matt Deeds, Nicole
  Hamilton, and Greg Hullender.
\newblock 2005.
\newblock Learning to rank using gradient descent.
\newblock In {\em Proceedings of the 22nd international conference on Machine
  learning}, pages 89--96. ACM.

\bibitem[\protect\citename{Burges}2010]{burges2010ranknet}
Christopher~JC Burges.
\newblock 2010.
\newblock From ranknet to lambdarank to lambdamart: An overview.
\newblock {\em Learning}, 11:23--581.

\bibitem[\protect\citename{Cao \bgroup et al.\egroup }2007]{cao2007learning}
Zhe Cao, Tao Qin, Tie-Yan Liu, Ming-Feng Tsai, and Hang Li.
\newblock 2007.
\newblock Learning to rank: from pairwise approach to listwise approach.
\newblock In {\em Proceedings of the 24th international conference on Machine
  learning}, pages 129--136. ACM.

\bibitem[\protect\citename{Das \bgroup et al.\egroup }2014]{das2014frame}
Dipanjan Das, Desai Chen, Andr{\'e}~FT Martins, Nathan Schneider, and Noah~A
  Smith.
\newblock 2014.
\newblock Frame-semantic parsing.
\newblock {\em Computational Linguistics}, 40(1):9--56.

\bibitem[\protect\citename{Freund \bgroup et al.\egroup
  }2003]{freund2003efficient}
Yoav Freund, Raj Iyer, Robert~E Schapire, and Yoram Singer.
\newblock 2003.
\newblock An efficient boosting algorithm for combining preferences.
\newblock {\em The Journal of machine learning research}, 4:933--969.

\bibitem[\protect\citename{Hang}2011]{hang2011short}
LI~Hang.
\newblock 2011.
\newblock A short introduction to learning to rank.
\newblock {\em IEICE TRANSACTIONS on Information and Systems},
  94(10):1854--1862.

\bibitem[\protect\citename{Herbrich \bgroup et al.\egroup
  }1999]{herbrich1999large}
Ralf Herbrich, Thore Graepel, and Klaus Obermayer.
\newblock 1999.
\newblock Large margin rank boundaries for ordinal regression.
\newblock {\em Advances in neural information processing systems}, pages
  115--132.

\bibitem[\protect\citename{Hermann \bgroup et al.\egroup
  }2015]{hermann2015teaching}
Karl~Moritz Hermann, Tomas Kocisky, Edward Grefenstette, Lasse Espeholt, Will
  Kay, Mustafa Suleyman, and Phil Blunsom.
\newblock 2015.
\newblock Teaching machines to read and comprehend.
\newblock In {\em Advances in Neural Information Processing Systems}, pages
  1684--1692.

\bibitem[\protect\citename{Kusner \bgroup et al.\egroup }2015]{kusner2015word}
Matt Kusner, Yu~Sun, Nicholas Kolkin, and Kilian~Q Weinberger.
\newblock 2015.
\newblock From word embeddings to document distances.
\newblock In {\em Proceedings of the 32nd International Conference on Machine
  Learning (ICML-15)}, pages 957--966.

\bibitem[\protect\citename{Mikolov \bgroup et al.\egroup
  }2013]{mikolov2013distributed}
Tomas Mikolov, Ilya Sutskever, Kai Chen, Greg~S Corrado, and Jeff Dean.
\newblock 2013.
\newblock Distributed representations of words and phrases and their
  compositionality.
\newblock In {\em Advances in neural information processing systems}, pages
  3111--3119.

\bibitem[\protect\citename{Narasimhan and Barzilay}2015]{narasimhan2015machine}
Karthik Narasimhan and Regina Barzilay.
\newblock 2015.
\newblock Machine comprehension with discourse relations.
\newblock In {\em 53rd Annual Meeting of the Association for Computational
  Linguistics}.

\bibitem[\protect\citename{Richardson \bgroup et al.\egroup
  }2013]{richardson2013mctest}
Matthew Richardson, Christopher~JC Burges, and Erin Renshaw.
\newblock 2013.
\newblock Mctest: A challenge dataset for the open-domain machine comprehension
  of text.
\newblock In {\em EMNLP}, volume~1, page~2.

\bibitem[\protect\citename{Sachan \bgroup et al.\egroup
  }2015]{sachan2015learning}
Mrinmaya Sachan, Avinava Dubey, Eric~P Xing, and Matthew Richardson.
\newblock 2015.
\newblock Learning answerentailing structures for machine comprehension.
\newblock In {\em Proceedings of ACL}.

\bibitem[\protect\citename{Smith \bgroup et al.\egroup }2015]{smithstrong}
Ellery Smith, Nicola Greco, Matko Bo{\v{s}}njak, and Andreas Vlachos.
\newblock 2015.
\newblock A strong lexical matching method for the machine comprehension test.

\bibitem[\protect\citename{Taylor}1953]{taylor1953cloze}
Wilson~L Taylor.
\newblock 1953.
\newblock Cloze procedure: a new tool for measuring readability.
\newblock {\em Journalism and Mass Communication Quarterly}, 30(4):415.

\bibitem[\protect\citename{Trischler \bgroup et al.\egroup
  }2016]{trischler2016parallel}
Adam Trischler, Zheng Ye, Xingdi Yuan, Jing He, Phillip Bachman, and Kaheer
  Suleman.
\newblock 2016.
\newblock A parallel-hierarchical model for machine comprehension on sparse
  data.
\newblock {\em arXiv preprint arXiv:1603.08884}.

\bibitem[\protect\citename{Wang and McAllester}2015]{wang2015machine}
Hai Wang and Mohit Bansal Kevin Gimpel~David McAllester.
\newblock 2015.
\newblock Machine comprehension with syntax, frames, and semantics.
\newblock {\em Volume 2: Short Papers}, page 700.

\bibitem[\protect\citename{Xu and Li}2007]{xu2007adarank}
Jun Xu and Hang Li.
\newblock 2007.
\newblock Adarank: a boosting algorithm for information retrieval.
\newblock In {\em Proceedings of the 30th annual international ACM SIGIR
  conference on Research and development in information retrieval}, pages
  391--398. ACM.

\end{thebibliography}
\bibliographystyle{emnlp2016}

\end{document}